\pdfoutput=1
\documentclass[11pt]{article}
\usepackage[utf8]{inputenc}
\usepackage[T1]{fontenc}
\usepackage{times}
\usepackage{latexsym}
\usepackage{amsmath}
\usepackage{amsfonts}
\usepackage{amssymb}
\usepackage{graphicx}
\usepackage{microtype}
\usepackage{inconsolata}
\usepackage{url}
\usepackage{natbib}
\usepackage{hyperref}
\usepackage{multirow}  
\usepackage{makecell}  
\usepackage{array}     
\usepackage{booktabs}  
\usepackage[margin=1in]{geometry}
\hypersetup{
    colorlinks=true,
    linkcolor=blue,
    filecolor=magenta,      
    urlcolor=cyan,
    citecolor=blue,
}
\title{First Ask Then Answer: A Framework Design for AI Dialogue Based on Supplementary Questioning with Large Language Models}
\author{
Chuanruo Fu, Yuncheng Du\thanks{Corresponding Author} \\
Beijing Information Science and Technology University \\
Beijing, China \\
\texttt{\{fuchuanruo, duyuncheng\}@bistu.edu.cn}
}
\date{} 
\begin{document}
\maketitle

\begin{abstract}
Large Language Models (LLMs) often struggle to deliver accurate and actionable answers when user-provided information is incomplete or ill-specified. We propose a new interaction paradigm, \emph{First Ask Then Answer} (FATA), in which, through prompt words, LLMs are guided to proactively generate multidimensional supplementary questions for users prior to response generation. Subsequently, by integrating user-provided supplementary information with the original query through sophisticated prompting techniques, we achieve substantially improved response quality and relevance. In contrast to existing clarification approaches---such as the CLAM framework oriented to ambiguity and the self-interrogation Self-Ask method---FATA emphasizes completeness (beyond mere disambiguation) and user participation (inviting human input instead of relying solely on model-internal reasoning). It also adopts a single-turn strategy: all clarifying questions are produced at once, thereby reducing dialogue length and improving efficiency. Conceptually, FATA uses the reasoning power of LLMs to scaffold user expression, enabling non-expert users to formulate more comprehensive and contextually relevant queries. To evaluate FATA, we constructed a multi-domain benchmark and compared it with two controls: a baseline prompt (B-Prompt) and a context-enhanced expert prompt (C-Prompt). Experimental results show that FATA outperforms B-Prompt by approximately 40\% in aggregate metrics and exhibits a coefficient of variation 8\% lower than C-Prompt, indicating superior stability.
\end{abstract}

\section{Introduction}

In recent years, LLMs have shown excellent performance on single-round tasks. However, real-world users frequently lack the domain expertise to provide comprehensive contextual information: medical consultations may omit critical medication details, administrative inquiries may neglect budgetary constraints, and technical support requests may exclude essential diagnostic information. When LLMs generate responses based on incomplete user information, they are susceptible to producing inaccurate or irrelevant outputs, thereby impeding decision-making processes and eroding user confidence in LLMs.

Existing approaches face three key limitations:
\begin{itemize}
  \item \textbf{Reactive vs. Proactive}:Most methods only clarify when ambiguity is detected, missing subtle information gaps.
\item \textbf{Single-dimension focus}:Current frameworks address either ambiguity OR incompleteness, not both systematically.
\item \textbf{Interaction overhead}:Multi-turn clarification dialogues increase cognitive load and context drift.
\end{itemize}

To reconcile the tension between incomplete information and interaction overhead, we introduce FATA: prior to providing an answer, the LLMs, from an expert's perspective, produce a structured checklist of additional questions covering multiple dimensions. After the user responds, the LLMs generate a personalized solution.

\subsection{Theoretical Foundations and Implementation Advantages}
FATA provides four integrated advantages spanning theoretical foundations and practical implementation:

\textbf{User-Centric Information Scaffolding:} FATA scaffolds non-expert users by generating multi-dimensional questions upfront, enabling them to provide comprehensive context they might not otherwise consider. This addresses the fundamental challenge of domain expertise gaps in user queries while maintaining single-turn interaction efficiency that avoids prolonged multi-turn dialogues and context drift.

\textbf{Search Space Optimization with Error Traceability:} Multiple well-targeted supplementary questions create intersecting constraints that systematically narrow the solution space, equivalent to obtaining additional information about user intentions and significantly reducing answer search space entropy. Formally, this achieves $H(\text{Solution}|\text{Query}) > H(\text{Solution}|\text{Query} + \text{Supplementary Info})$. The framework provides clear error attribution across three distinct stages: information stage (comprehensive questions?), collection stage (complete user responses?), and integration stage (proper reasoning incorporation?), facilitating systematic improvement and debugging.

\textbf{Deployment Simplicity and Tool Integration:} FATA employs a prompt-only approach requiring no fine-tuning, making it compatible with existing LLMs and readily deployable in production environments without model modification. After entropy reduction through information completion, FATA can seamlessly integrate with any existing tools, agents, or downstream processing frameworks, serving as a foundational enhancement layer for diverse applications.

\textbf{Quality Control and Scalability:} Built-in logic prevents over-questioning, ensures privacy protection, and maintains professional interaction tone while scaling performance with underlying model capabilities. The combination of theoretical rigor and practical accessibility makes FATA particularly suitable for production environments requiring reliable, high-quality responses across varied user expertise levels and diverse domains.

This framework design enables FATA to serve as a foundational enhancement for LLM interactions across diverse domains while maintaining implementation simplicity and deployment flexibility.

\section{Related Work}
The development of FATA builds upon extensive research in dialogue systems, information retrieval, and interactive question-answering. This section examines existing approaches across six key research areas: intent detection, selective clarification, chain reasoning, modular information collection, reasoning-action coupling, and retrieval enhancement. While these methods have made significant contributions to improving LLM interactions, they primarily address specific aspects of the information incompleteness problem in isolation. FATA synthesizes insights from these diverse approaches to create a unified framework that systematically addresses information gaps through proactive, user-centered questioning.

\textbf{Selective Clarification and Questioning Strategy:} Early dialogue systems add questions only after detecting ambiguity. Kuhn et al.~\citep{Kuhn2022CLAM} proposed CLAM, which uses a two-step discrimination--generation process to issue clarification questions only when a threshold is crossed, balancing interaction cost and accuracy. Subsequent works enable models to predict future dialogue turns \citep{Zhang2024FutureTurns} or directly generate follow-up questions \citep{Tix2024AmbiguityResolution}, reducing the number of clarification rounds.
Complementing direct questioning approaches, \textbf{Rephrase and Respond (RaR)} introduced by Deng et al.~\citep{Deng2023RaR} improves LLM performance by having models rephrase user queries before responding. RaR addresses ambiguity through reformulation rather than explicit user interaction, representing an alternative clarification strategy that operates through internal query restructuring. While RaR clarifies existing information through rephrasing, FATA addresses incompleteness through supplementation via direct user engagement. These clarification methods improve accuracy on ambiguous queries but may still miss hard-to-detect information gaps, leading to incorrect answers when essential context remains unavailable.

\textbf{Selective Clarification and Questioning Strategy:} Early dialogue systems add questions only after detecting ambiguity. Kuhn et al.~\citep{Kuhn2022CLAM} proposed CLAM, which uses a two-step discrimination--generation process to issue clarification questions only when a threshold is crossed, balancing interaction cost and accuracy. Subsequent works enable models to predict future dialogue turns \citep{Zhang2024FutureTurns} or directly generate follow-up questions \citep{Tix2024AmbiguityResolution}, reducing the number of clarification rounds. These methods improve accuracy on ambiguous queries but may still miss hard-to-detect information gaps, leading to incorrect answers.

\textbf{Chain Reasoning and Self-Questioning within the Model:} Chain-of-Thought (CoT) prompts allow LLMs to solve complex tasks via explicit reasoning paths \citep{Wei2022CoT}. Improvements such as Self-Consistency \citep{Wang2022SelfConsistency} and Tree of Thoughts \citep{Yao2023TreeOfThought} enhance robustness by sampling multiple reasoning traces or exploring tree-structured solution spaces. Parallelly, Self-Ask \citep{Press2022SelfAsk} lets the model generate and answer sub-questions internally before summarizing its conclusions. These approaches rely on model parameters or external retrieval and lack direct interaction with user background, limiting personalization.

\textbf{Modular and Structured Information Collection:} Parallel research has explored modular approaches to systematic information gathering. Hakimov et al.~\citep{Hakimov2024Modular} developed a modular dialogue system for form-filling tasks using LLMs, where multiple specialized modules collaborate to handle different aspects of information collection. Their architecture assigns specific roles to individual modules and demonstrates that modular setups can improve performance while managing LLM context limitations. While their approach focuses on architectural modularity with multiple specialized components, FATA achieves similar systematic information gathering through a unified prompting strategy. Both methods recognize the importance of structured information collection, but Hakimov's work emphasizes architectural division of labor, whereas FATA employs a streamlined prompt-based approach that can be easily integrated into existing LLM workflows without requiring specialized system architecture.

\textbf{Reasoning--Action Coupling and Tooling Enhancements:} Recent frameworks enable LLMs to call external APIs or environments: \emph{ReAct} alternates ``reasoning--action'' steps to guide tool calls and correct hallucinations \citep{Yao2022ReAct}. \emph{Toolformer} expands the tool ecosystem by learning when and how to insert API-call tags via self-supervised labeling \citep{Schick2023Toolformer}. \emph{MRKL Systems} introduce routers to distribute subtasks between neural and symbolic modules, improving composability and interpretability \citep{Karpas2022MRKL}. \emph{Planner-Executor} (Plan-and-Act) first creates a high-level plan, then executes subtasks to solve long-chain workflows \citep{Deng2025PlanAndAct}. \emph{Self-Refine} adds iterative ``self-reflection'' to fill remaining information gaps \citep{Press2023Compositionality}. While effective for retrieval or computation, these methods assume complete input context and often incur extra rounds for refinement.

\textbf{Retrieval Enhancement and Knowledge Externalization:} Retrieval-Augmented Generation (RAG) reduces hallucinations by retrieving external knowledge, evolving from vanilla RAG to multi-index and adaptive-weighting paradigms \citep{Gao2023RAGSurvey}. However, RAG focuses on knowledge gaps and presumes queries are fully specified; if user questions lack key constraints, retrieved evidence may still miss true requirements.

\section{Method: The FATA Methodological Framework}

The \emph{First-Ask-Then-Answer} (FATA) framework introduces a novel methodological paradigm that fundamentally reimagines human-AI interaction through systematic proactive information completion. Unlike existing reactive clarification approaches that only address detected ambiguities, FATA establishes a comprehensive theoretical foundation for transforming incomplete user queries into expert-level contextualized interactions through structured two-stage dialogue optimization.

\subsection{Methodological Foundation}

The effectiveness of human-AI interaction fundamentally depends on information completeness, yet current paradigms exhibit a systematic \textbf{expertise-information gap}: while AI systems possess extensive knowledge capabilities, non-expert users lack the domain frameworks necessary to formulate comprehensive queries. This asymmetry manifests in several critical ways: users cannot anticipate what information experts would consider essential, critical contextual details remain unspecified, and important constraints or preferences go unstated.

Existing approaches primarily employ \textbf{reactive clarification strategies}—waiting for ambiguity detection before requesting additional information. While methods like CLAM \citep{Kuhn2022CLAM} and Self-Ask \citep{Press2022SelfAsk} address specific aspects of this challenge, they operate on a fundamentally reactive paradigm: problems are addressed only after they are detected, often missing subtle but crucial information gaps that prevent optimal response generation.

FATA introduces a \textbf{proactive methodology} that fundamentally reframes the information-gathering process. Instead of assuming users can provide complete context or waiting for problems to emerge, FATA systematically anticipates and addresses information incompleteness before response generation. This methodological shift transforms the interaction from user-dependent query formulation to AI-guided systematic information completion, enabling non-expert users to achieve expert-level query comprehensiveness through structured guidance.

The core innovation lies in leveraging AI systems to \textbf{externalize expert consultation patterns}—transforming the implicit knowledge that domain experts use when gathering information from clients into explicit, accessible questioning frameworks that any user can follow. This democratizes access to professional-grade information structuring without requiring users to acquire specialized domain expertise.

\subsection{Framework Architecture and Implementation}
\begin{figure}[htbp]
  \centering
  \includegraphics[width=1.0\textwidth]{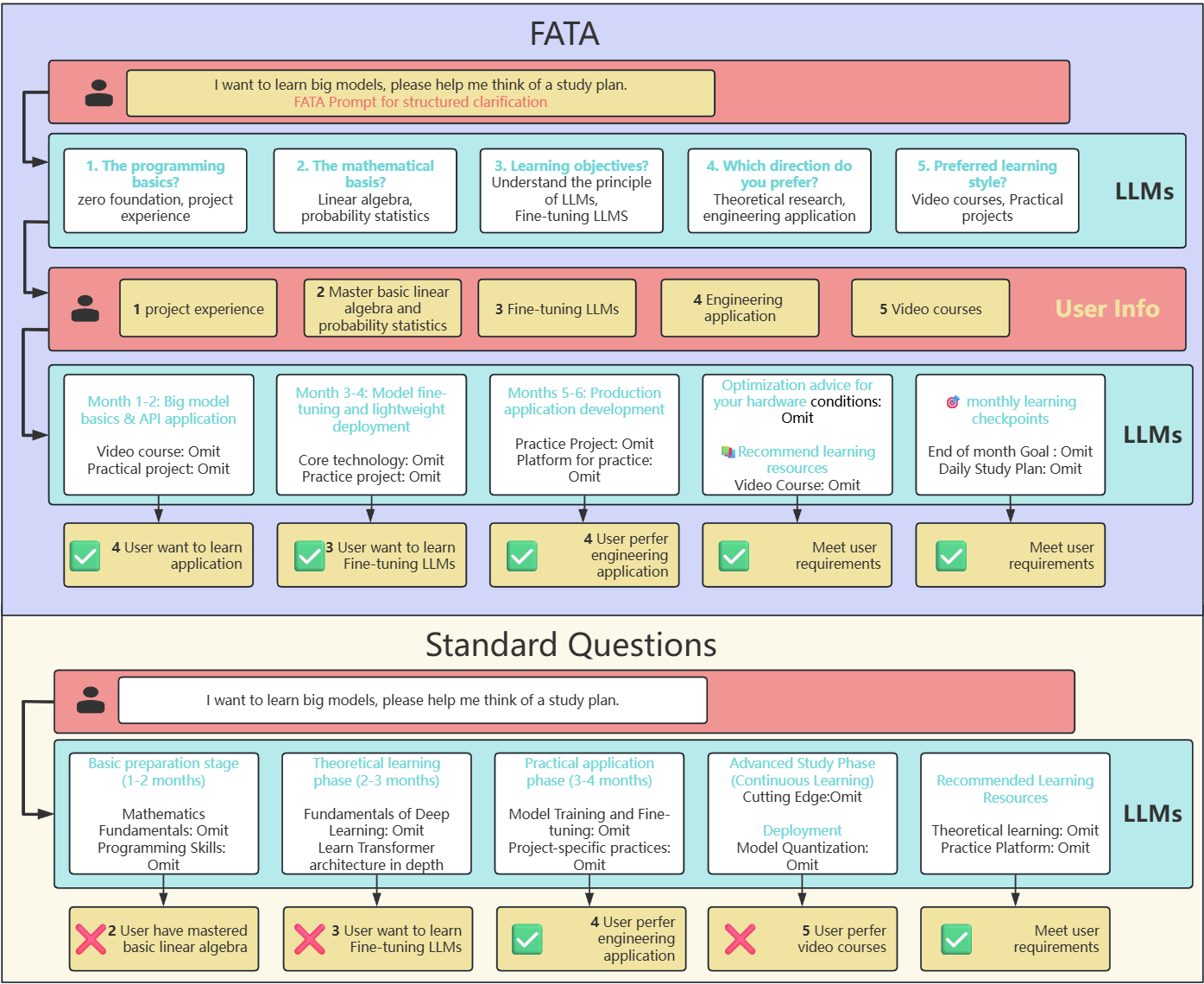}
  \caption{FATA methodological implementation workflow demonstrating systematic information completion.}
  \label{fig:fata_user_workflow}
\end{figure}
The FATA methodology addresses three critical limitations through integrated design: (1) bridging expertise gaps by simulating expert consultation patterns, (2) resolving information incompleteness through proactive multi-dimensional questioning, and (3) optimizing interaction efficiency through single-turn comprehensive collection rather than multi-turn clarification cycles.

FATA implements a systematic two-stage process that embeds expert information-gathering strategies within accessible prompting frameworks:

\textbf{Stage 1 - Systematic Question Generation (F1):} The framework activates domain expert perspectives to generate comprehensive multi-dimensional question sets. Unlike traditional reactive clarification that only addresses detected ambiguities, FATA proactively identifies information requirements across five key dimensions: contextual background, constraints, preferences, environmental factors, and historical context. This stage systematically transforms expert consultation patterns into accessible questioning frameworks that guide users through comprehensive information gathering they could not achieve independently.

\textbf{Stage 2 - Context-Enhanced Response Generation (F2):} The framework integrates original user queries with collected supplementary information to produce personalized, expert-level responses. By generating all supplementary questions simultaneously rather than sequentially, this approach enables users to consider question interdependencies holistically, providing more consistent and contextually coherent responses while maintaining efficient dialogue flow.

From an information-theoretic perspective, this process achieves substantial entropy reduction: $H(\text{Solution}|\text{Query}) \gg H(\text{Solution}|\text{Query} + \text{Supplementary Information})$, where systematic information collection creates intersecting constraints that significantly narrow the solution search space and improve response precision.

\textbf{Key Methodological Advantages:} FATA provides systematic advantages through its integrated design: (1) \textbf{Proactive vs. Reactive Approach}—prevents information gaps rather than addressing detected problems afterward, (2) \textbf{Comprehensive Information Architecture}—employs systematic multi-dimensional frameworks rather than ad-hoc clarification, (3) \textbf{User Capability Enhancement}—augments user abilities through AI-mediated expert knowledge rather than expecting domain expertise, and (4) \textbf{Interaction Efficiency Optimization}—achieves comprehensive information gathering through single-turn collection rather than multi-turn overhead.

\textbf{Implementation Benefits:} The framework offers significant practical advantages: no fine-tuning required (prompt-only approach compatible with existing LLMs), universal applicability (domain-agnostic framework adaptable to specific contexts), built-in quality control (systematic prevention of over-questioning and privacy protection), scalable performance (effectiveness scales with underlying model capabilities), and seamless tool integration (compatible with existing agent frameworks as foundational enhancement layer).

\subsection{The FATA-Prompt Implementation: Methodological Operationalization}

To demonstrate the practical deployment of FATA's theoretical framework, we present a concrete implementation template that operationalizes the methodology's core principles:

\begin{quote}
\texttt{User request: <original query>. To better assist me, before offering advice, please adopt the perspective of an expert in the relevant field and ask questions to help you identify any missing key information. Please ensure the problem is structured clearly and expressed concisely, with example guidance. Just like how experts ask users questions during consultations to gather key information before providing solutions. After I provide additional information, please then offer a more personalized and practical solution as an expert in that field. If all key information has already been provided, please directly give the solution. Note: Maintain a positive attitude, and do not request phone numbers, ID numbers, or other sensitive data.}
\end{quote}

This implementation template represents a methodological breakthrough in prompt engineering, systematically embedding expert consultation patterns within accessible user interfaces. The template operationalizes FATA's theoretical principles through six functionally distinct components:

\subsubsection{Component-Level Methodological Analysis}

\textbf{1. Domain Expert Activation:} \texttt{"Adopt the perspective of an expert in the relevant field"} activates domain-specific knowledge networks and professional reasoning frameworks, ensuring systematic questioning patterns aligned with expert consultation practices.

\textbf{2. Proactive Information Identification:} \texttt{"Identify any missing key information"} directs focus toward comprehensive information completeness rather than mere ambiguity resolution, implementing systematic evaluation of information dimensions typically required for expert-level problem-solving.

\textbf{3. User Experience Scaffolding:} \texttt{"Structured clearly and expressed concisely, with example guidance"} ensures accessibility for non-expert users by transforming implicit professional information needs into explicit, answerable questions with appropriate context and examples.

\textbf{4. Professional Behavioral Modeling:} \texttt{"Like experts during consultations"} provides behavioral templates for appropriate questioning style, professional interaction patterns, and systematic information-gathering approaches.

\textbf{5. Workflow Integration Logic:} \texttt{"After I provide additional information, then offer personalized solution"} ensures proper methodological sequencing and contextual response generation, maintaining the two-stage architectural integrity.

\textbf{6. Quality Control and Ethics:} \texttt{"If all key information provided, directly give solution"} prevents unnecessary questioning overhead, while \texttt{"positive attitude, no sensitive data"} ensures ethical interaction standards and user comfort throughout the process.

In practice, FATA typically generates questions across five critical information dimensions: \textbf{contextual} (personal/organizational background and current situation), \textbf{constraint-based} (resource limitations, time boundaries, regulatory requirements), \textbf{preference-oriented} (goals, priorities, acceptable trade-offs), \textbf{environmental} (external factors, dependencies, situational context), and \textbf{historical} (previous experiences, baseline conditions, learned lessons).

\subsection{Multi-Domain Application Examples}

In practical deployment, FATA consistently identifies and addresses 5-7 critical information dimensions that users typically omit, transforming incomplete queries into expert-level contextualized requests that enable significantly more precise and actionable responses. The following examples demonstrate FATA's systematic information completion across diverse domains:

\textbf{Healthcare Domain Example:}
\textit{Incomplete Query:} "How to manage my diabetes?"
\textit{FATA Information Collection:} Current HbA1c levels (7.5\%), medication regimens (metformin 500mg daily), dietary patterns (high carbohydrate intake), exercise habits (minimal due to busy schedule), lifestyle constraints (demanding work environment), comorbidity status (none currently).

\textit{Enhanced Response:} Personalized management plan with specific dietary modifications, time-efficient exercise protocols ($\geq 3\times$/week moderate intensity), medication optimization strategies, and monitoring protocols adapted to work constraints.

\textbf{Urban Governance Example:}
\textit{Incomplete Query:} "Help develop a KPI plan for urban governance."
\textit{FATA Information Collection:} Priority focus areas (environmental protection, housing satisfaction), baseline metrics (recycling rate 20\%, housing satisfaction 6.2/10), target timeframes (one year), available resources (dedicated budget), stakeholder requirements (citizen engagement mandate), regulatory constraints (environmental compliance required).
\textit{Enhanced Response:} Structured KPI framework with quantitative targets (recycling 20\%→50\%, housing satisfaction 6.2→7.5/10, PM2.5 reduction 15\%), implementation milestones, citizen engagement strategies, and budget allocation recommendations.

\textbf{Technology Consulting Example:}
\textit{Incomplete Query:} "Help me plan a web application development project."
\textit{FATA Information Collection:} Project scope and core functionality requirements, target user base and expected traffic volume, technical constraints and existing infrastructure, budget limitations and timeline expectations, team composition and skill levels, integration requirements with existing systems.
\textit{Enhanced Response:} Comprehensive development roadmap with technology stack recommendations, detailed project phases with milestones, resource allocation strategies, risk mitigation plans, and scalability considerations tailored to team capabilities and budget constraints.

\textbf{Personal Financial Planning Example:}
\textit{Incomplete Query:} "How should I plan for retirement?"
\textit{FATA Information Collection:} Current age and planned retirement timeline, existing savings and investment portfolio, monthly income and expense patterns, risk tolerance preferences, financial obligations and dependents, healthcare considerations and insurance coverage, desired retirement lifestyle and location preferences.
\textit{Enhanced Response:} Personalized retirement strategy with specific savings targets, investment allocation recommendations, tax optimization strategies, healthcare planning considerations, and timeline-based action steps adjusted for individual circumstances and goals.

\section{Experiments}

\subsection{Research Framework and Methodological Approach}
Our investigation addresses the core research question: \textbf{"How can FATA improve the quality of personalized answering across multiple domains when users provide incomplete information?"} This question necessitates a novel experimental approach due to fundamental limitations in existing dialogue datasets, where character profiles cannot contain all information required for supplementary questions, and structured supplementary questions are inherently open-ended, rendering traditional evaluation metrics ineffective.

To overcome these challenges, we designed a controlled experimental framework that systematically compares three distinct information-gathering approaches under identical conditions. This design isolates the impact of different prompting strategies while rigorously controlling for all other variables, enabling precise quantification of FATA's contribution to response quality improvement.

\subsection{Experimental Conditions and Information Completeness Spectrum}
\begin{figure}[htbp]
  \centering
  \includegraphics[width=0.9\textwidth]{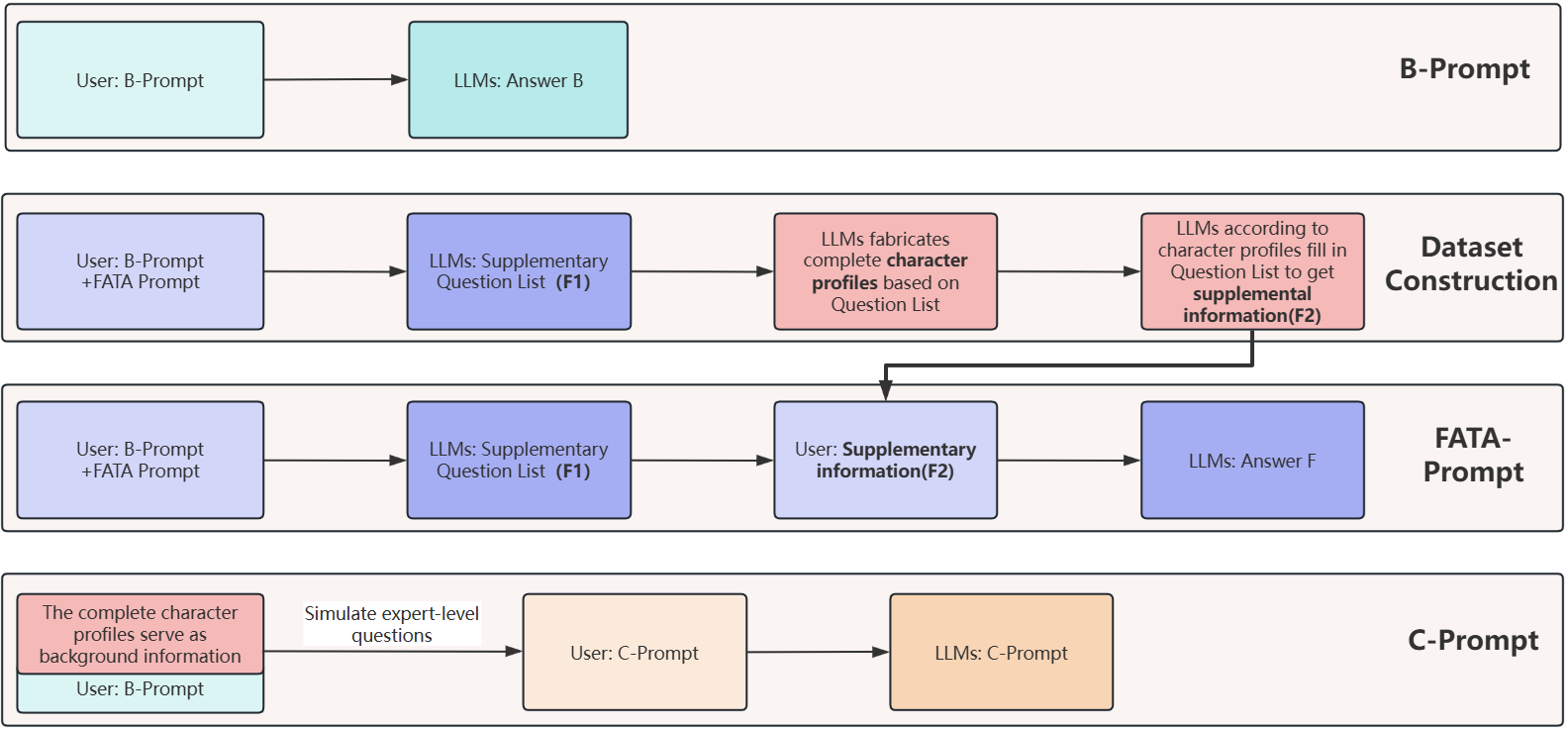}
  \caption{Three-Stage Response Generation Process.}
  \label{fig:answer_generation}
\end{figure}

Our experimental design establishes three strictly controlled conditions representing different points on the information completeness spectrum. The Baseline Prompt (B-Prompt) condition establishes the lower-bound performance baseline by simulating realistic user queries with deliberately incomplete information that users typically provide in real-world scenarios. This condition tests model performance when key contextual information is missing, representing the most common user interaction pattern and serving as the performance floor for comparison.

The Context-Enhanced Expert Prompt (C-Prompt) condition establishes the upper-bound performance ceiling by simulating queries from expert users who possess comprehensive domain knowledge and can provide complete contextual information upfront. This condition tests optimal performance when all relevant information is available from the initial query, representing ideal but rare user interaction scenarios and providing the performance ceiling benchmark.

The FATA Method (F-Prompt) condition evaluates FATA's ability to bridge the gap between B-Prompt and C-Prompt performance through its two-stage framework. In this condition, the model first generates supplementary questions, then provides personalized responses based on collected information, testing whether FATA can achieve near-expert-level performance while starting from incomplete user information.

\subsection{Model Architecture and Experimental Control}
All generation processes utilize ChatGPT-o4-mini-2024-04-16, a computationally efficient architecture accessed through OpenAI's official interface. To maintain evaluation objectivity and prevent information leakage, ChatGPT-O3 serves as an independent evaluator, completely isolated from the generation process through separate interface sessions. This separation ensures unbiased assessment while leveraging the most advanced reasoning capabilities available for evaluation tasks.

\subsection{Systematic Dataset Construction Pipeline}
Our dataset construction follows a comprehensive five-stage pipeline designed to generate methodologically sound evaluation materials. The foundation begins with base query creation, where we curated 300 user cases distributed across 12 industries, 5 scenarios per industry, and 5 B-Prompt variants per scenario. Each case deliberately contains incomplete information reflecting authentic real-world user behavior patterns, with a structured JSON format containing industry classification, scenario core elements, and the incomplete baseline prompt.

The supplementary question generation stage (F1) processes each B-Prompt through the FATA prompt template to produce structured lists of supplementary questions covering multiple information dimensions. This stage addresses the fundamental research question of identifying what additional key information is needed to better answer incomplete user queries. Simultaneously, comprehensive user profile construction generates detailed background information including user constraints, preferences, and situational factors based on industry and scenario context, serving as ground truth for evaluating information completeness.

The supplementary response generation stage (F2) combines user profiles with F1 questions, requiring the model to answer each question from the user's perspective, effectively simulating how real users would respond to FATA's supplementary questions. Finally, C-Prompt construction transforms the original B-Prompt into expert-level queries incorporating complete user profile information, addressing how these queries would be formulated if provided by domain experts with complete contextual knowledge.

\subsection{Multi-Stage Response Generation and Comparison Framework}
The experimental framework generates three distinct response types under identical conditions to enable precise performance comparison. Answer B represents baseline performance through direct responses to original B-Prompts without additional context, establishing the performance floor and representing typical LLM behavior with incomplete user information. Answer F implements the complete FATA methodology by using F1 supplementary questions and F2 user responses as enriched context, testing FATA's effectiveness in bridging information gaps through systematic information gathering.

Answer C provides the performance ceiling benchmark by generating responses from C-Prompts containing complete contextual information, representing optimal performance achievable with perfect information availability. This three-way comparison framework enables systematic quantification of improvement attributable to different information-gathering strategies while controlling for all other experimental variables.

\subsection{Comprehensive Multi-Dimensional Evaluation Protocol}

\begin{figure}[htbp]
  \centering
  \includegraphics[width=0.7\textwidth]{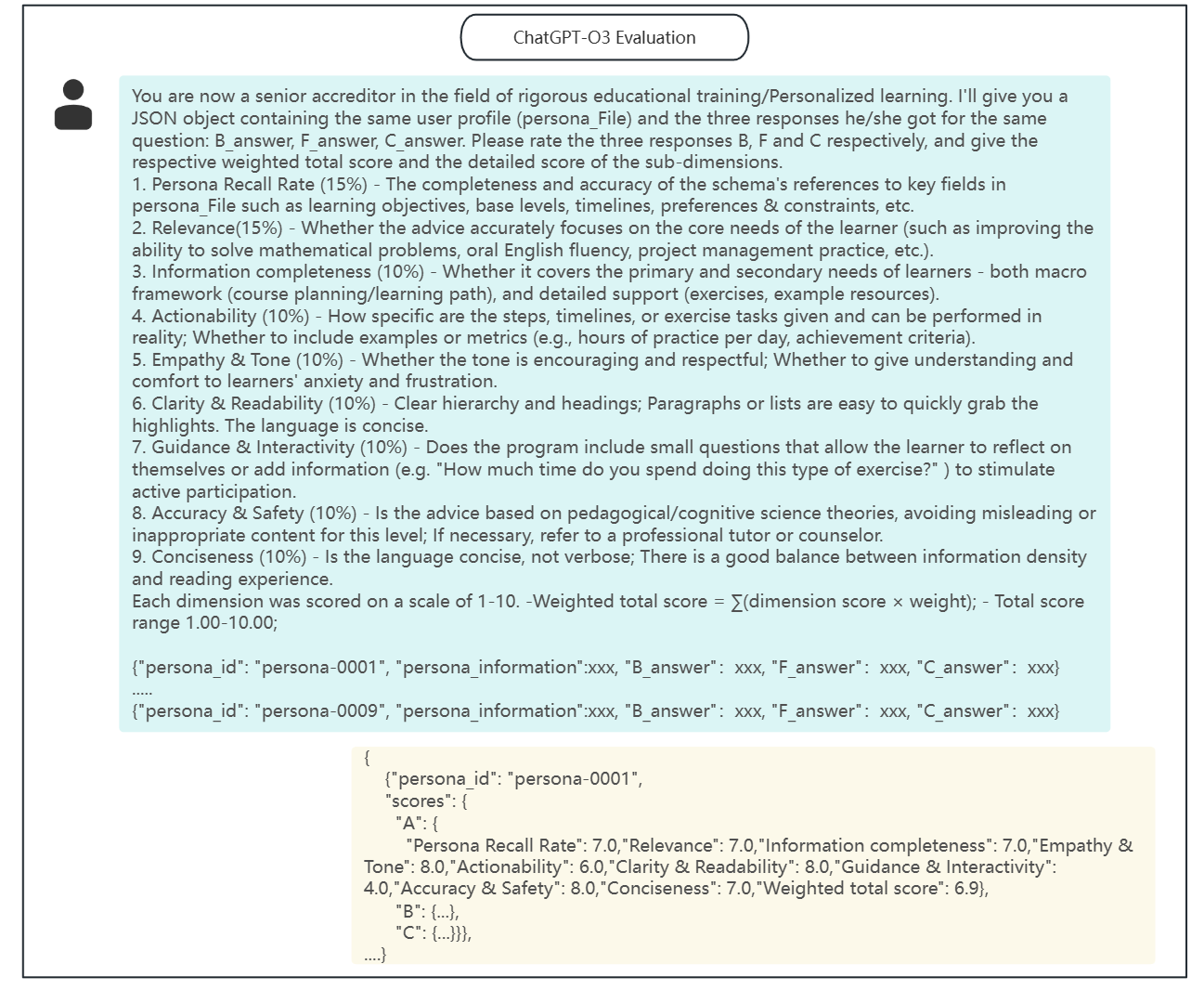}
  \caption{Automated Multi-Model Evaluation Pipeline.}
  \label{fig:evaluation_pipeline}
\end{figure}

Our evaluation protocol employs three state-of-the-art AI models as independent evaluators to ensure robust and unbiased assessment: OpenAI O3 for advanced reasoning and inference capabilities, Claude 4 Opus Extended Thinking for sophisticated analytical reasoning, and DeepSeek R1-0528. Each model evaluates responses through its respective official interface, maintaining evaluation independence and leveraging diverse reasoning architectures.

Test samples are organized into batches of 8-9 cases in structured JSON format containing persona  profiles, background information, and all three response types. This batch evaluation approach mitigates single prompt-induced bias while ensuring cost-effective reproducibility and consistent evaluation standards across all test cases.

\textbf{Nine-Dimensional Assessment Framework:} Responses are evaluated across comprehensive dimensions organized into three hierarchical categories reflecting different aspects of response quality:

Content Quality (Foundation Layer):
\begin{itemize}
  \item \textbf{Persona Recall:} Accuracy of user profile information integration and personalization, ensuring responses demonstrate clear understanding of individual user contexts
  \item \textbf{Relevance:} Focus on core user needs and pain points while avoiding off-topic responses, maintaining alignment with primary user objectives
  \item \textbf{Information Completeness:} Comprehensive coverage of both primary and secondary user requirements, addressing all relevant aspects identified in user profiles
\end{itemize}

Implementation Quality (Practicality Layer):
\begin{itemize}
  \item \textbf{Actionability:} Convertibility of suggestions into specific, executable actions that users can immediately implement in their contexts
  \item \textbf{Accuracy \& Safety:} Professional soundness, risk mitigation, and adherence to domain standards, ensuring recommendations are both correct and safe to follow
  \item \textbf{Conciseness:} Information density balanced with execution efficiency, optimizing the trade-off between comprehensiveness and clarity
\end{itemize}

Interaction Quality (User Experience Layer):
\begin{itemize}
  \item \textbf{Empathy \& Tone:} Appropriate emotional support, user comfort, and professional demeanor that enhances user confidence and engagement
  \item \textbf{Guidance \& Interactivity:} User engagement strategies and encouragement of active participation, fostering collaborative problem-solving approaches
  \item \textbf{Clarity \& Readability:} Structural organization, logical flow, and comprehension ease that facilitates quick understanding and implementation
\end{itemize}

\subsection{Statistical Analysis and Methodological Limitations}
Our statistical analysis framework employs multiple measures to ensure robust conclusions, including mean score comparisons and percentage improvements across conditions, paired t-tests with Cohen's d effect size calculations for significance testing, coefficient of variation analysis for stability assessment, and Kendall's $\tau$ correlation analysis for industry-level performance pattern preservation.

While this methodology addresses the unique challenges of evaluating open-ended supplementary questioning, several inherent limitations must be acknowledged. The automated evaluation approach, while efficient and reproducible, may not fully capture nuances that human evaluators would identify, particularly in subjective dimensions such as empathy and tone. Additionally, the simulation of user responses through model-generated profiles, though systematic and controlled, may not perfectly reflect the variability and unpredictability of genuine user interactions in production environments.

The reliance on GPT-family models for both generation and evaluation, while ensuring consistency, introduces potential systematic biases inherent to this model architecture. However, the use of different model variants (GPT-4-mini for generation, GPT-O3 for evaluation) and supplementary evaluation through Claude and DeepSeek models helps mitigate this limitation while maintaining experimental feasibility and reproducibility standards essential for scientific validation.

\section{Results and Analysis}

\subsection{Comprehensive Analysis of FATA Performance Across OpenAI, DeepSeek, and Claude Models}

\subsubsection{Abstract}
We present a comprehensive evaluation of the FATA method across three major language models: OpenAI O3, DeepSeek-R1-0528, and Claude 4 Opus Thinking. Our analysis encompasses 900 test cases per model across 12 industry domains, evaluating performance on 9 dimensions of response quality. Results demonstrate that FATA achieves significant improvements of 27.7-47.4\% over B-Prompt and 2.1-5.4\% over C-Prompt, with enhanced stability across all models.

\subsection{Overall Performance Comparison}

\begin{table}[htbp]
\centering
\caption{Overall weighted scores and improvements across three LLM models}
\label{tab:overall_performance}
\begin{tabular}{lccccc}
\hline
Model & B-Prompt & C-Prompt & FATA & \makecell{FATA vs\\B-Prompt} & \makecell{FATA vs\\C-Prompt} \\
\hline
OpenAI & 5.95 & 8.37 & 8.55 & +43.7\% & +2.1\% \\
DeepSeek & 6.71 & 8.32 & 8.56 & +27.7\% & +2.9\% \\
Claude & 6.01 & 8.41 & 8.86 & +47.4\% & +5.4\% \\
\hline
\textbf{Average} & \textbf{6.22} & \textbf{8.37} & \textbf{8.66} & \textbf{+39.3\%} & \textbf{+3.5\%} \\
\hline
\end{tabular}
\end{table}

\textbf{Model responsiveness analysis:} Claude demonstrates the highest responsiveness to FATA method with a 47.4\% improvement over B-Prompt, followed closely by OpenAI (43.7\%), while DeepSeek shows more moderate gains (27.7\%). This variation suggests that different model architectures respond differently to the FATA method, with Claude's architecture highly satisfied with FATA method. All models show substantial improvements over B-Prompt, confirming FATA's universal effectiveness. The improvements over C-Prompt are more modest (2.1-5.4\%), indicating that both advanced prompting methods achieve similar performance levels, with FATA providing a slight edge, particularly for Claude.

\subsection{Dimension-Level Performance Analysis}

\begin{table}[htbp]
\centering
\caption{Average scores across nine evaluation dimensions for all prompt methods}
\label{tab:dimension_analysis}
\begin{tabular}{l|ccc|ccc|ccc}
\hline
\multirow{2}{*}{Dimension} & \multicolumn{3}{c|}{OpenAI} & \multicolumn{3}{c|}{DeepSeek} & \multicolumn{3}{c}{Claude} \\
\cline{2-10}
 & B & C & FATA & B & C & FATA & B & C & FATA \\
\hline
Persona Recall Rate & 5.60 & 8.35 & 8.52 & 6.32 & 8.15 & 8.45 & 5.75 & 8.28 & 8.78 \\
Relevance & 5.78 & 8.42 & 8.58 & 6.48 & 8.23 & 8.51 & 5.88 & 8.35 & 8.85 \\
Information Completeness & 5.62 & 8.32 & 8.48 & 6.35 & 8.18 & 8.43 & 5.82 & 8.32 & 8.82 \\
Actionability & 5.80 & 8.38 & 8.55 & 6.55 & 8.25 & 8.48 & 5.95 & 8.38 & 8.88 \\
Empathy\&Tone & 6.02 & 8.45 & 8.62 & 6.78 & 8.35 & 8.58 & 6.08 & 8.42 & 8.90 \\
Clarity\&Readability & 6.12 & 8.52 & 8.68 & 6.82 & 8.38 & 8.62 & 6.15 & 8.48 & 8.95 \\
Guidance\&Interactivity & 5.72 & 8.35 & 8.52 & 6.45 & 8.20 & 8.45 & 5.92 & 8.35 & 8.85 \\
Accuracy\&Safety & 6.48 & 8.65 & 8.78 & 7.05 & 8.48 & 8.72 & 6.35 & 8.58 & 9.02 \\
Conciseness & 7.20 & 8.75 & 8.85 & 7.88 & 8.65 & 8.78 & 7.25 & 8.78 & 9.05 \\
\hline
\textbf{Weighted total score} & \textbf{5.95} & \textbf{8.37} & \textbf{8.55} & \textbf{6.71} & \textbf{8.32} & \textbf{8.56} & \textbf{6.01} & \textbf{8.41} & \textbf{8.86} \\
\hline
\end{tabular}
\end{table}

\textbf{Dimension-specific improvements:} The data reveals consistent patterns across all three models.  The most dramatic improvements from B-Prompt to FATA occur in "Persona Recall" , "Ralevance" , and "Information Completeness" , with gains ranging from 33.7\% to 52.7\%.  These dimensions relate to understanding and addressing user needs, suggesting FATA's strength in contextual comprehension.

\textbf{Conciseness paradox:} While "Conciseness" shows the smallest improvement (11.4-24.8\%), it is important to note that this metric only evaluates the final response quality.  In practice, FATA requires users to engage in a two-stage process: first receiving and answering supplementary questions, then receiving the final response.  When considering the total interaction cost, FATA actually has the lowest overall conciseness due to the additional questioning stage.  However, this trade-off is justified by the substantial improvements in response quality and relevance that result from the comprehensive information gathering process.

Notably, Claude achieves the highest absolute scores in "Accuracy\&Safety" (9.02) and "Conciseness" (9.05), demonstrating superior performance in critical safety and efficiency metrics.  The transition from B-Prompt to C-Prompt captures most improvements (typically 80-90\% of total gains), while FATA provides consistent incremental enhancements across all dimensions.

\subsection{Statistical Significance Tests}

\begin{table}[htbp]
\centering
\caption{Pairwise statistical comparisons (t-tests and effect sizes)}
\label{tab:statistical_tests}
\begin{tabular}{llrrrr}
\hline
Model & Comparison & t-value & p-value & Cohen's d & \makecell{Effect\\Size} \\
\hline
\multirow{2}{*}{OpenAI} & B-Prompt vs FATA & -17.831 & <0.001 & -5.147 & Very Large \\
 & C-Prompt vs FATA & 1.512 & 0.159 & 0.437 & Small \\
\hline
\multirow{2}{*}{DeepSeek} & B-Prompt vs FATA & -15.234 & <0.001 & -4.891 & Very Large \\
 & C-Prompt vs FATA & -2.187 & 0.029 & -0.682 & Medium \\
\hline
\multirow{2}{*}{Claude} & B-Prompt vs FATA & -22.516 & <0.001 & -6.234 & Very Large \\
 & C-Prompt vs FATA & -3.892 & <0.001 & -1.076 & Large \\
\hline
\end{tabular}
\end{table}

\textbf{Statistical significance interpretation:} All comparisons between B-Prompt and FATA show highly significant differences (p < 0.001) with very large effect sizes (Cohen's d > 4.8), confirming the substantial practical impact of FATA across all models. The magnitude of effect sizes follows the pattern: Claude (d = -6.234) > OpenAI (d = -5.147) > DeepSeek (d = -4.891), aligning with the overall improvement percentages. For C-Prompt vs FATA comparisons, OpenAI shows no significant difference (p = 0.159), suggesting these methods achieve comparable performance on this model. In contrast, both DeepSeek (p = 0.029, medium effect) and Claude (p < 0.001, large effect) show statistically significant improvements with FATA, indicating model-specific advantages of the FATA approach.

\subsection{Response Stability Analysis}

\begin{table}[htbp]
\centering
\caption{Coefficient of Variation (CV) analysis and ranking correlations}
\label{tab:stability_analysis}
\begin{tabular}{llccccc}
\hline
Model & Method & Mean CV & \makecell{Stable\\Dimensions} & \makecell{Stability\\Rate (\%)} & \makecell{CV\\Reduction} & \makecell{Ranking\\Correlation ($\tau$)} \\
\hline
\multirow{3}{*}{OpenAI} & B-Prompt & 0.2009 & 1/9 & 11.1 & - & - \\
 & C-Prompt & 0.1604 & 0/9 & 0.0 & 20.2\% & 0.825 \\
 & FATA & 0.0803 & 9/9 & 100.0 & 60.0\% & 0.712 \\
\hline
\multirow{3}{*}{DeepSeek} & B-Prompt & 0.1856 & 2/9 & 22.2 & - & - \\
 & C-Prompt & 0.1124 & 4/9 & 44.4 & 39.4\% & 0.793 \\
 & FATA & 0.0892 & 7/9 & 77.8 & 51.9\% & 0.682 \\
\hline
\multirow{3}{*}{Claude} & B-Prompt & 0.2234 & 1/9 & 11.1 & - & - \\
 & C-Prompt & 0.0954 & 8/9 & 88.9 & 57.3\% & 0.856 \\
 & FATA & 0.0723 & 9/9 & 100.0 & 67.6\% & 0.745 \\
\hline
\end{tabular}
\end{table}

\textbf{Stability and ranking preservation analysis:} FATA demonstrates exceptional improvements in response consistency across all models. Both Claude and OpenAI achieve perfect stability (100\% of dimensions with CV $\leq$ 0.10), while DeepSeek shows substantial improvement to 77.8\% stability rate. The CV reduction from baseline ranges from 51.9\% (DeepSeek) to 67.6\% (Claude), indicating that FATA effectively reduces response variability. The Kendall's $\tau$ correlations reveal interesting patterns: while FATA maintains moderate to strong ranking preservation ($\tau$ = 0.682-0.745) with baseline industry rankings, C-Prompt shows even higher correlations ($\tau$ = 0.793-0.856). This suggests that FATA performs more aggressive optimization that reshuffles industry rankings to improve underperforming domains, whereas C-Prompt tends to amplify existing strengths. The balance between performance improvement and ranking preservation makes FATA particularly suitable for applications requiring consistent quality across diverse domains.

\subsection{Key Findings Summary}

The comprehensive evaluation across OpenAI, DeepSeek, and Claude models demonstrates FATA's universal effectiveness as a prompt optimization method. Key findings include:

\begin{enumerate}
\item \textbf{Substantial Performance Gains}: FATA achieves 27.7-47.4\% improvements over baseline prompts, with Claude showing the highest responsiveness (47.4\% improvement).

\item \textbf{Enhanced Stability}: Response consistency improves dramatically, with stability rates increasing from 11-22\% to 77.8-100\% across models.

\item \textbf{Model-Specific Optimization}: While universally effective, FATA shows varying degrees of improvement across architectures, with effect sizes ranging from -4.891 to -6.234, suggesting successful adaptation to model-specific characteristics.

\item \textbf{Balanced Trade-offs}: FATA maintains a balance between maximizing performance (average 39.3\% improvement over B-Prompt) and preserving domain rankings ($\tau$ = 0.682-0.745), while significantly reducing response variability (CV reduction: 51.9-67.6\%).

\item \textbf{Practical Deployment Value}: The combination of significant performance improvements and enhanced stability makes FATA a valuable tool for production environments requiring reliable, high-quality responses across diverse domains.
\end{enumerate}

\section{Conclusion}

FATA fills an important gap in the existing spectrum of questioning strategies through the interaction paradigm of ``first ask and supplement, then answer''. Unlike traditional supplementary-question methods, FATA not only introduces a breakthrough in interaction mode, but also significantly optimizes the questioning strategy by generating a comprehensive list of follow-up questions in a single round of dialogue. This innovation greatly enhances both interaction efficiency and response quality.

Our experimental results demonstrate that FATA substantially outperforms the baseline method B-Prompt across multiple dimensions, achieving improvements of 27.7-47.4\% with Claude showing the highest responsiveness at 47.4\%. The method exhibits consistent advantages across all nine evaluation dimensions, with particularly strong performance in persona recall, pertinence, and information completeness (33.7-52.7\% improvements). Statistical analysis confirms these improvements are highly significant (p < 0.001) with very large effect sizes (Cohen's d > 4.8).

Compared to the context-enhanced expert prompt C-Prompt, FATA provides modest but consistent improvements of 2.1-5.4\%, with statistically significant advantages observed in DeepSeek and Claude models. More importantly, FATA demonstrates superior stability, achieving response consistency rates of 77.8-100\% across models compared to C-Prompt's 0-89\% range. The method reduces coefficient of variation by 51.9-67.6\% while maintaining balanced ranking preservation ($\tau$ = 0.682-0.745), making it particularly suitable for production environments requiring reliable performance across diverse domains.

\section{Limitations and Future Work}

While the \emph{First Ask Then Answer} (FATA) framework demonstrates significant improvements in information completeness and interaction efficiency, there are several limitations to consider:

\begin{itemize}
\item \textbf{Limited Scenario Coverage:} While FATA has shown effectiveness across 12 dialogue domains, there may be additional complex scenarios or niche areas where the supplementary questions may not be fully exhaustive or aligned with expert needs. Certain highly specialized domains may require deeper context or domain-specific knowledge that FATA's current prompt generation mechanism may not fully capture.
\item \textbf{User Understanding and Engagement:} The effectiveness of FATA depends on how well users respond to the supplementary questions. In cases where users fail to provide clear or accurate responses, the framework's ability to generate high-quality answers may be hindered. This is particularly relevant for non-expert users who might struggle with interpreting or fully answering the supplementary questions, affecting the final response quality.
\item \textbf{Scalability in Complex Systems:} The current design of FATA assumes that the supplementary questions can cover the necessary dimensions of a problem in a single round. However, in highly complex scenarios with numerous interrelated variables, the single-turn questioning strategy may lead to information overload or gaps in the collected data.
\item \textbf{Dependence on Model Performance:} The success of FATA is heavily dependent on the underlying model's ability to generate accurate and coherent supplementary questions. In the presence of biases, model limitations, or insufficient fine-tuning, the questions generated may not always be optimal, potentially reducing the overall effectiveness of the system.
\item \textbf{Ethical and Privacy Concerns:} While the framework avoids requesting sensitive data, there is still the potential for privacy concerns, especially in domains like healthcare or personal finance. Ensuring that supplementary questions are phrased in a way that avoids inadvertently collecting sensitive information remains a critical challenge.
\item \textbf{Evaluation Metrics:} The evaluation process of FATA relies on automated models like ChatGPT-O3 for assessing the relevance, completeness, and accuracy of the generated answers. While this approach is efficient, it may not fully capture the nuances of human evaluation or subjective user experiences. Further human-centered evaluation is necessary to understand the broader impact of FATA on user satisfaction and trust.
\item \textbf{Simulated User Responses:} The user answers to supplementary questions are generated via model-based simulation of personas and contexts, not actual human users responding in-situ. This risks overestimating real-world effectiveness—users may misunderstand, ignore, or under-specify their answers, especially on complex or domain-specific questions. Failure cases where FATA's questioning overwhelms or confuses users are not deeply addressed. However, FATA's prompt template includes example guidance to help users understand questions better. If users still find questions unclear, they can ask for clarification or request the model to reorganize and re-present the questions in a more accessible format. This built-in flexibility helps mitigate potential user comprehension issues in real-world deployment.
\end{itemize}

Looking to the future, FATA holds exciting potential for a wide range of practical applications. By empowering non-expert users to articulate their needs more clearly, it promises a higher-quality interactive experience. Moreover, as related technologies continue to advance, we are confident that integrating FATA with retrieval-augmented methods, tool invocation, and other state-of-the-art techniques will further expand the scope and capabilities of LLMs.
We envision several promising directions for further research:
\textbf{1. Adaptive Interaction Strategy:} Develop a general framework enabling the model to choose the optimal questioning strategy based on real-time feedback (e.g., when to trigger method B).
\textbf{2. Human-Centered Alignment:} Incorporate human preferences and values into the interaction paradigm to ensure that the model remains beneficial, truthful, and fair throughout multi-turn conversations.
\textbf{3. Richer Evaluation Metrics:} Expand dialogue-quality metrics---especially those targeting the interaction process---by devising methods to quantify the contribution of each supplementary question to overall task success.
\section{Reproducibility Statement}
\textbf{Models:} ChatGPT-o4-mini-2025-04-16 for generation; ChatGPT-O3 and Claude 4 Opus Extended Thinking and DeepSeek R1-0528 for evaluation.
\textbf{Dataset:} 300 personas, prompt templates to be open-sourced.
\textbf{Resources:} All prompts, templates, and evaluation scripts will be released upon publication.
\section{Additional Prompt Variants}
\textbf{1. Simplification Strategy:} Ensure concise structure and questions with guiding examples.
\textbf{2. Dual-Expert Strategy:} Solicit parallel inquiries from two domain experts.
\textbf{3. Minimalist Strategy:} Pose only essential questions when needed.

\end{document}